\documentclass[10pt,twocolumn,letterpaper]{article}

\usepackage{verbatim}
\usepackage{iccv}
\usepackage{times}
\usepackage{epsfig}
\usepackage{graphicx}
\usepackage{amsmath}
\usepackage{amssymb}
\usepackage{subfig}
\usepackage{multirow}

\usepackage[pagebackref=true,breaklinks=true,letterpaper=true,colorlinks,bookmarks=false]{hyperref}

 \iccvfinalcopy 

\ificcvfinal\fi
\begin{document}


\title{Deep Learning for Face Recognition: Pride or Prejudiced? }


\author{Shruti Nagpal, Maneet Singh, Richa Singh, Mayank Vatsa\\
IIIT-Delhi, India\\
\tt\small \{shrutin, maneets, rsingh, mayank\}@iiitd.ac.in
}

\maketitle

\begin{abstract}
Do very high accuracies of deep networks suggest pride of effective AI or are deep networks prejudiced? Do they suffer from in-group biases (own-race-bias and own-age-bias), and mimic the human behaviour? Is in-group specific information being encoded sub-consciously by the deep networks? 

\noindent This research attempts to answer these questions and presents an in-depth analysis of `bias' in deep learning based face recognition systems. This is the first work which decodes if and where bias is encoded for face recognition. Taking cues from cognitive studies, we inspect if deep networks are also affected by social in- and out-group effect. Networks are analyzed for own-race and own-age bias, both of which have been well established in human beings. The sub-conscious behavior of face recognition models is examined to understand if they encode race or age specific features for face recognition. Analysis is performed based on 36 experiments conducted on multiple datasets. Four deep learning networks either trained from scratch or pre-trained on over 10M images are used. Variations across class activation maps and feature visualizations provide novel insights into the functioning of deep learning systems, suggesting behavior similar to humans. It is our belief that a better understanding of state-of-the-art deep learning networks would enable researchers to address the given challenge of bias in AI, and develop fairer systems.
\end{abstract}

\section{Introduction}
Deep learning based systems show extremely high accuracy in many computer vision application such as object detection and recognition, and image synthesis \cite{tinyface, lfw, hyperface}. In particular, significantly better results have been shown in many face analytic tasks including face detection, age and gender prediction, and face recognition leading many researchers to claim \textit{super human} performance in face recognition. However, recently reported bias in face analytics shows the issues with the current state of the art deep learning face analytic systems \cite{aies}.

\begin{figure}
     \centering
     \subfloat[][(L-R): Recent incident of bias in AI (image taken from the internet); own-race bias already established in humans] {\includegraphics[width=3.2in]{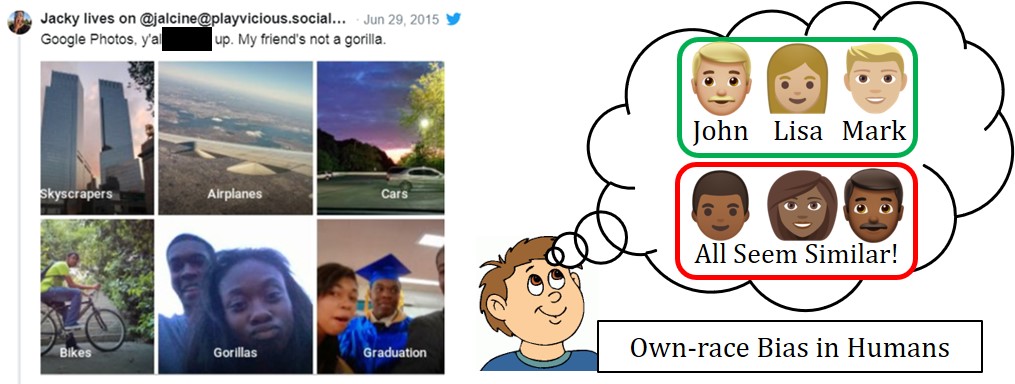}}
     \\\vspace{-5pt}
     \subfloat[][Pride or Prejudiced? Own-race bias observed in deep learning models]{\includegraphics[clip, trim=0.1cm 0cm 0.25cm 0cm, width=3.2in] {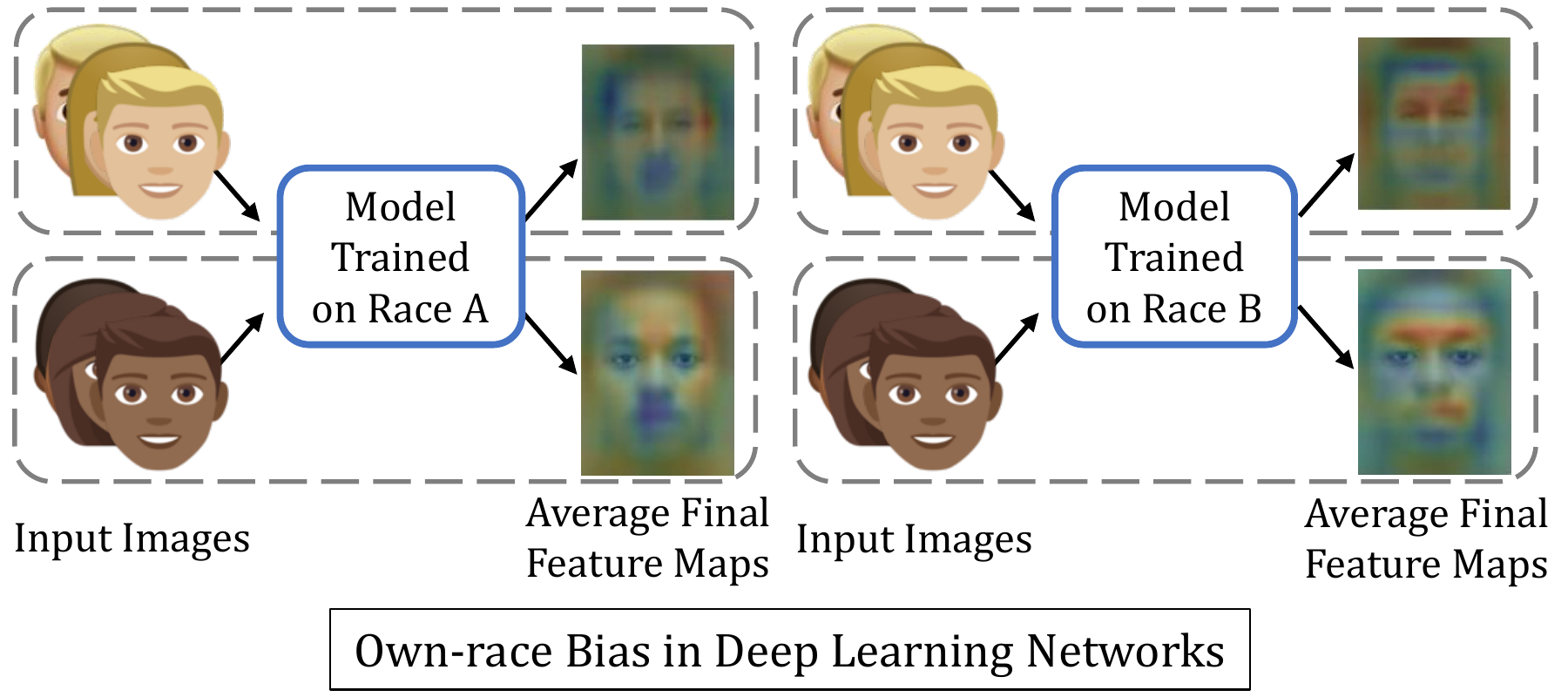}} 
     \caption{Bias observed in (a) AI systems and humans, and (b) deep learning networks obtained via this research. Current state-of-the-art deep learning models appear to mimic the human in-group biases by focusing on different facial regions when trained on specific sub-groups for face recognition. Best viewed in color.}
     \label{fig:intro}
     \vspace{-15pt}
\end{figure}

In July 2018, Amazon's facial recognition tool, `Rekognition', falsely identified 28 members of the US Congress as criminals \cite{amazon}. The US Congress contains around 20\% members of color, out of which 40\% were mis-classified. The study, conducted by the American Civil Liberties Union (ACLU), is one of the recent incidents shedding light on the emerging \textit{biased} behavior of Artificial Intelligence systems, and the need to develop \textit{fairer AI}. Earlier examples include the ProPublica study of algorithms being used by US courts and parole boards to predict future behaviour of criminals \cite{proPublica}. It was observed that the system was biased against black defendants, and gave them a higher score as compared to their white counterparts. Another popular instance which shook the Machine Learning community was of the Google Photos application's mis-classification of a black couple as gorilla (Figure \ref{fig:intro}(a)) \cite{Google}. Multiple repeated incidents suggesting existence of bias demand dedicated research in order to ensure the development of AI systems for the \textit{better}, and not for the \textit{worse}. This research focuses on understanding the hidden biases of \textit{face recognition systems}, specifically, current state-of-the-art deep learning networks. The existence of bias with respect to the race and age of an individual has been analyzed in face recognition models, by answering \textit{if} and \textit{where} the bias is encoded. 

Bias is often defined as an \textit{inclination or prejudice for or against one person or group, especially in a way considered to be unfair}. While bias in AI is a relatively new field, humans have shown to possess biases across different parameters since the very beginning. 
Neuro-cognitive researchers have long established the presence of \textit{in-group} biases\footnote{In-group bias refers to the tendency to have an affinity towards members belonging to the same group as the individual.}  in humans, wherein we tend to recognize individuals of our race or age easily, as compared to individuals of an \textit{out-group} \cite{biasGroups}. One such instance dates back to 1971, when five black men were acquitted for the murder of Khomas Revels \cite{story}. Despite the lack of any forensic evidence, five eyewitnesses testified and identified them as offenders. However, during the third trial, private investigators located the actual culprits, who were later convicted. Interestingly, all the five men initially acquitted were black and all the five eyewitnesses were white. Social Psychologist, Dr. William Haythorn identified the reason for mis-identifications to be the cross-racial identifications, because of which people of the other race look alike, resulting in biased outcomes (Figure \ref{fig:intro}(a)). Several studies later, this phenomenon was termed as the \textit{own race bias}, or \textit{other race} or \textit{cross-race effect}, where people of the other race look alike. 

We believe that the human trait of making biased decisions has also made its way into the recent AI systems for computer vision (Figure \ref{fig:intro}(b)). Since most of these systems today are based on deep learning, they provide limited insights into the decision making process. Thus, understanding the models and their corresponding biases is of utmost importance, in order to prevent the inclusion of biases that existed in the past into our future. 
\subsection{Literature Review: Bias in Facial Analysis} 
A majority of research has focused on understanding bias in models developed on textual data \cite{Bolukbasi16Nips}, with limited attention to computer vision tasks \cite{anne2018women}. However, recent research has focused on understanding the presence of bias in commercial-off-the-shelf systems and deep learning models primarily for gender and ethnicity classification in face images \cite{genderShades,das18mitigating,ryu18inclusive} and \textbf{not} face recognition. Buolamwini and Gebru \cite{genderShades} categorized face images into four categories based on their skin color: darker males, darker females, lighter males, and lighter females. The performance of three commercial gender classification systems was analyzed with respect to the proposed categorization. It was observed that, out of the four categories, dark-skinned females were the most mis-classified group, thereby creating a need for fairer unbiased facial analysis algorithms. It is important to note that the authors analyzed gender classification using three commercial systems, without exploring the performance of deep learning models. 

The establishment of bias for tasks such as gender classification, led to the requirement of \textit{fairer} algorithms, capable of performing a particular task, unbiased of other related covariates. Das \textit{et al.} \cite{das18mitigating} proposed a Multi-Task Convolutional Neural Network (MTCNN) for joint gender, race, and age classification of face images. The proposed model demonstrates improved performance for the given tasks across multiple subgroups of gender, age, and ethnicity. Simultaneously, Alvi \textit{et al.} \cite{alvi18blind} proposed a joint learning and unlearning (JLU) technique for eliminating bias from CNN networks. Experiments are performed for multiple tasks of age, gender, race, and pose classification of face images, independently, while the other covariates are used for inducing bias. Ryu \textit{et al.} \cite{ryu18inclusive} proposed using transfer learning to develop attribute prediction models inclusive of different race and gender subgroups via a novel InclusiveFaceNet model. It demonstrates improved performance of attribute classification across different subgroups.  

\subsection{Research Contributions}
This research extends beyond the existing literature by analyzing deep learning based \textit{face recognition models} for bias existence. This is the first of its kind research which attempts to understand what deep networks are encoding, and whether they learn features similar to the human brain, with respect to different co-variates of face recognition. Four deep learning networks (LightCNN-9 \cite{lightcnn}, LightCNN-29 \cite{lightcnn}, ResNet50 \cite{resnet}, and SENet50 \cite{senet}), either trained from scratch or pre-trained on large scale datasets containing about 10M images have been evaluated to establish our findings. The key findings of this research are:
\begin{itemize}
\vspace{-6pt}
\item \textbf{Result-1:} Upon simulating the real world scenario of limited exposure to other races/age-groups by training on data belonging to a specific class, we observe that deep learning networks mimic the human tendency of \textit{in-group bias} across race and age. 
\vspace{-6pt}
\item \textbf{Result-2:} Similar to the human behavior of face recognition,  
race-specific \textit{regions of interest} are encoded in networks trained on a particular race.
\vspace{-6pt}
\item \textbf{Result-3:} Extending upon the existing cognitive literature, different regions of interest are also observed for varying age-groups, suggesting \textit{own-age bias} in deep learning networks. 
\vspace{-6pt}
\item \textbf{Result-4:} Networks pre-trained on large-scale data with varying distribution (eg. VGGFace2 \cite{vggface2} and MSCeleb-1M \cite{celeb} datasets), demonstrate high generalization abilities, however, bias across age persists to be a major challenge. 
\vspace{-6pt}
\item \textbf{Result-5:} Fine-tuning a pre-trained network results in change of \textit{region of interest}, which might reduce its generalization abilities.
\vspace{-6pt}
\end{itemize}
It is our belief that insights into the functioning of deep networks can help develop fairer AI systems, capable of unbiased decisions. Since deep learning models appear to mimic the human brain, findings of this research can also enable better utilization of pre-trained systems by applying suitable cognitive techniques (designed for humans) for de-biasing.
The remainder of this paper is organized as follows: the following section provides the experiments and protocols used for both the case studies of race and age. Sections \ref{sec:ethnicity} and \ref{sec:age} present the findings, along with the key takeaways and insights, followed by the conclusions of this research.   

\section{Experiments and Protocols}
In this research, deep-learning based face recognition networks are analyzed to answer two key questions: \vspace{-6pt}
\begin{itemize}
    \item  Does  deep  learning  encode  race-specific information? \vspace{-6pt}
    \item Does deep learning encode age-specific information? \vspace{-6pt}
\end{itemize}
We analyze the behavior of deep learning networks with respect to well established cognitive studies on humans, and identify the regions used by deep learning models to learn features for face recognition. We further delve deeper to observe if these regions are consistent across different races and age-groups, and how deep learning models behave when trained from scratch on selective data i.e. data highly biased towards specific subgroups. 

\noindent\textbf{Network Details:} Four deep learning based face recognition networks are analyzed to answer the two key questions pertaining to the effect and existence of bias due to race and age: (i) LightCNN-9 \cite{lightcnn}, (ii) Light CNN-29 \cite{lightcnn}, (iii) ResNet50 \cite{resnet}, and (iv) SENet50 \cite{senet}. LightCNN-9 is trained from scratch, while the remaining three networks are pre-trained on \textit{large-scale} face datasets. LightCNN-29 is pre-trained on the MS-Celeb-1M dataset\footnote{https://github.com/AlfredXiangWu/LightCNN} \cite{celeb}, which contains 10M images corresponding to nearly 10k identities in the training set. ResNet50 and SENet50 are pre-trained on the VGG-Face2 \cite{vggface2} and MS-Celeb-1M datasets for face recognition\footnote{https://github.com/cydonia999/VGGFace2-pytorch}. VGG-Face2 dataset contains 3.31M images pertaining to 9,131 identities. Therefore, the two models are trained on over 13M images, the largest corpus used for training a single face recognition model.

\noindent\textbf{Datasets Used for Case-study-1 (Effect of Race):} The behavior of deep learning networks is analyzed with respect to two races: \textit{Race-A (Caucasoid)} and \textit{Race-B (Congoid)}\footnote{For this study, we use the existing classification of races, namely Caucasoid and Congoid to understand machine learning based face recognition systems. The races are referred to as Race-A and Race-B in the manuscript.} 
Three datasets are used for performing the said analysis:

\noindent \textit{(i) CMU Multi-PIE} \cite{multipie}: Over 44K images of 336 subjects images are selected which correspond to frontal face images having illumination and expression variations. 

\noindent \textit{(ii) Craniofacial Longitudinal Morphological (MORPH) Album-2} \cite{morph}: Over 52k frontal face images pertaining to 10,409 subjects are used.

\noindent \textit{(iii) Racial Faces in the Wild (RFW)} \cite{rfw} is an unconstrained dataset of African (Race-B), Caucasian (Race-A), Asian, and Indian face images, collected from the Internet. Its test set contains labeled images \textit{across races}, with limited labeled Caucasian faces in the training set. For analysis, the test set containing 10,196 Race-A and 10,145 Race-B face images of 2,959 and 2,995 subjects, respectively, has been used.
    
\begin{figure}
\centering
\includegraphics[width=3in]{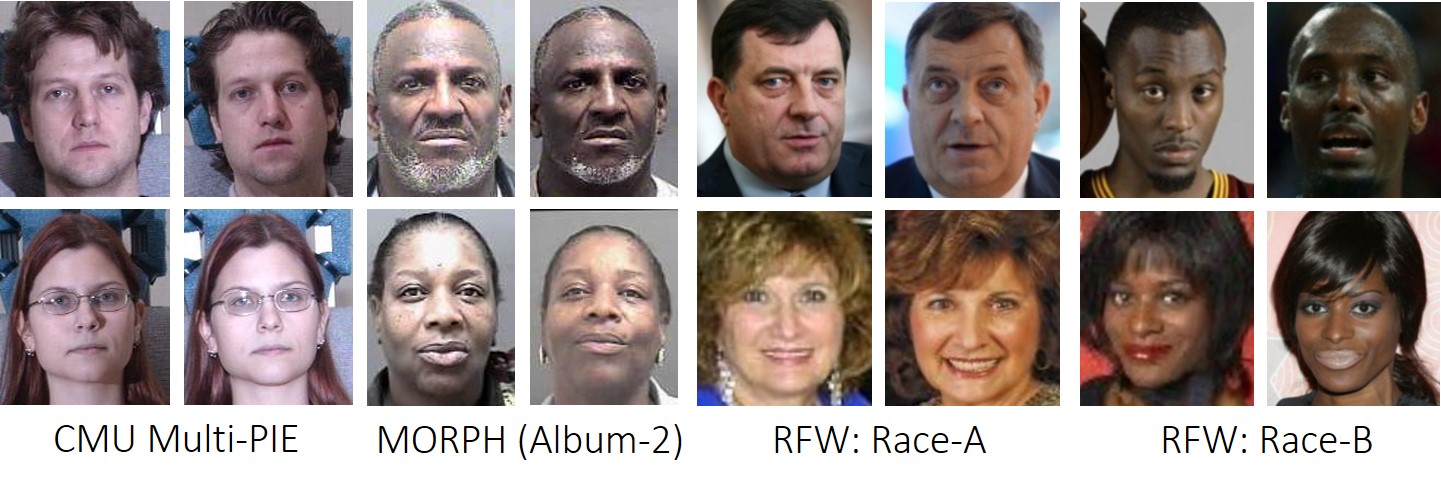}
\vspace{-10pt}
\caption{Samples from the datasets used for understanding racial bias in face recognition: (i) CMU Multi-PIE (Race-A), (ii) MORPH (Race-B), and (iii) RFW: Race-A and Race-B.}
\vspace{-10pt}
\label{fig:db}
\end{figure}

The CMU Multi-PIE dataset is used for analyzing the networks for Race-A, while the MORPH dataset is used for Race-B. Both the datasets are captured in constrained settings, and for both, 70\% of the subjects are used to create the training (or fine-tuning) partition, while the remaining 30\% subjects form the test set. This ensures disjoint training and testing partitions, in terms of images and subjects. Due to the availability of limited labeled training data in the RFW datset, only the test data is used to evaluate our hypothesis. Therefore, unless explicitly specified, analysis is drawn using the CMU Multi-PIE and MORPH datasets. Figure \ref{fig:db} presents sample images from the three datasets demonstrating facial images across the two races.

\begin{figure}
\centering
\includegraphics[width=3.2in]{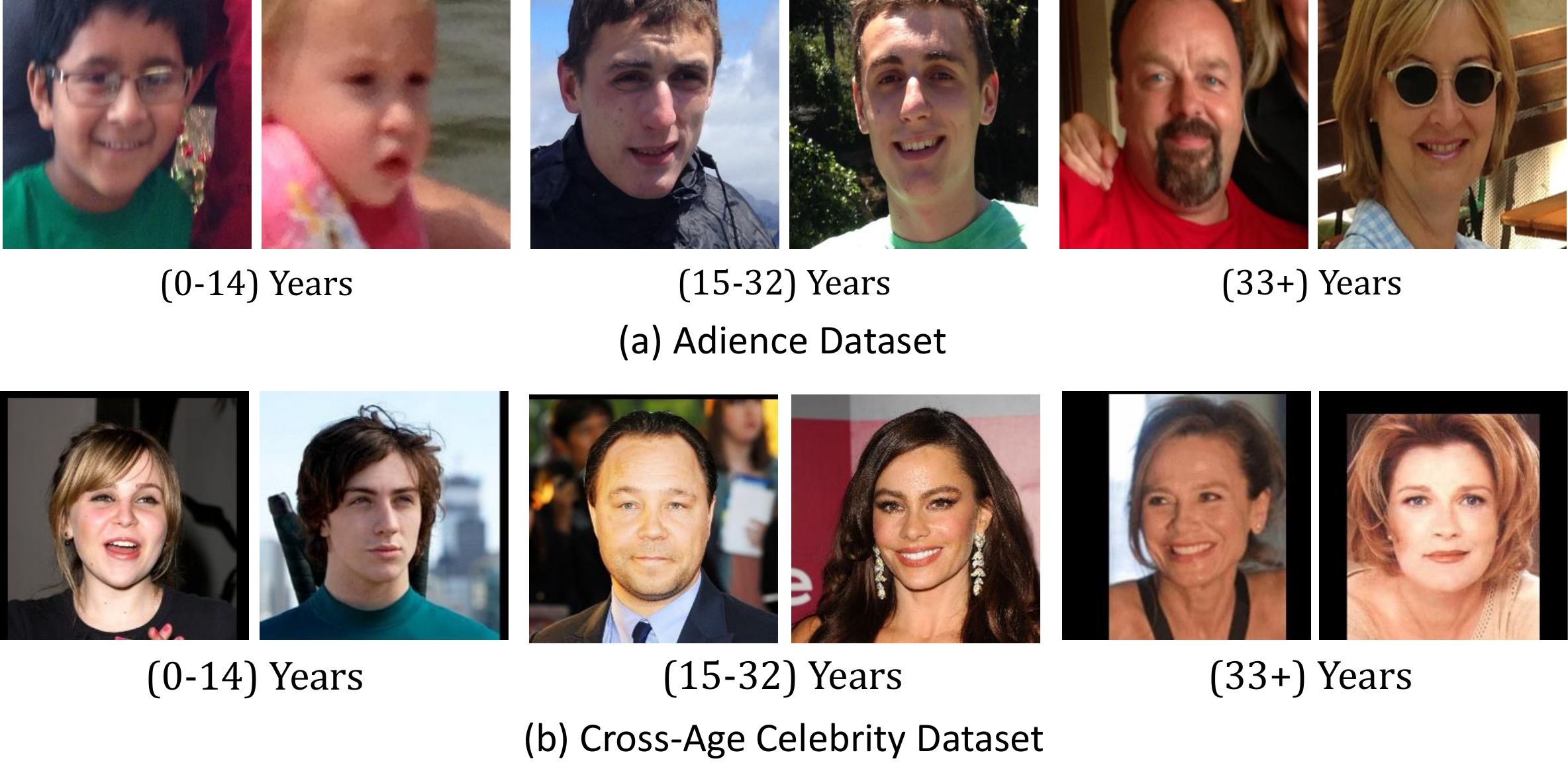}
\vspace{-10pt}
\caption{Sample images of the two datasets used for analyzing the effect of varying age on deep learning networks.}
\vspace{-10pt}
\label{fig:agedb}
\end{figure}

\noindent\textbf{Datasets Used for Case-study-2 (Effect of Age):} Deep learning networks are analyzed for face recognition with respect to three age groups: (i) (0-14) years, (ii) (15-32) years, and (iii) (33+) years. To the best of our knowledge, there does not exist any large scale dataset containing the identity and age information for such a large range. IMDB-Wiki \cite{imdb} is a large dataset containing face images with age and identity information, however, it has been observed that the dataset contains erroneous age information. Therefore, in this study we combined face images from two datasets: Adience dataset \cite{adience} and Cross-Age Celebrity Dataset (CACD) \cite{cacd}. Both the datasets contain images collected from the Internet, having pose, illumination, and background variations. Figure \ref{fig:agedb} presents sample images from both the datasets, displaying variations in age. The CACD dataset contains over 1,60,000 face images belonging to the age range of 14-62 years. In our experiments, the CACD dataset has been used for training (or fine-tuning) the deep learning networks. However, since it does not contain subjects below the age of 14 years, 70\% of the Adience dataset has also been used for training. Thus, the training or fine-tuning set is a combination of images from CACD and Adience datasets. The remaining 30\% of data pertaining to the Adience dataset is used for testing. Mutual exclusion (for both images and subjects) has been ensured between the training and testing partitions. 

\vspace{5pt}
\noindent\textbf{{Details Regarding Experiments and Analysis:}} Analysis is drawn from 36 experiments (Table \ref{tab:exp}) conducted across different case-studies, networks, and training data. At the time of training from scratch, models are trained on data pertaining to a single sub-group only (race or age), to observe the regions being learned for face recognition; and on equal number of samples belonging to different subgroups. Cross sub-group experiments have also been performed on the trained models to understand what regions are being used in each case to perform classification. For example, LightCNN-9 trained on Race-A subjects is used for evaluation on Race-B subjects, to study the difference in processing (if any). Three pre-trained networks ResNet50, SENet50, and LightCNN-29 have also been evaluated for performance and visualizations. The effect of fine-tuning has also been studied to understand how the regions of interest evolve for different sub-groups with pre-trained networks. Fine-tuning is performed on the entire network for the given train set. Feature-level analysis is performed with respect to the learned feature maps, where the final convolution layer maps are interpolated and super-imposed on the input image, to obtain the region of interest for the given input image and network. Class Activation Maps (CAMs) \cite{cam} are used to obtain the most discriminative regions of face images, focused upon by the CNNs. Results are also reported in terms of the verification accuracy (Genuine Acceptance Rate) obtained at 1\% False Acceptance Rate. In order to ensure consistency across analysis and fair comparisons, fixed protocols have been used for all experiments. Model files will be released for reproducibility of findings.

\begin{table}
\footnotesize       
\centering
\caption{List of 36 experiments performed to understand face recognition in deep learning models w.r.t. race and age bias. Results have been shown using three datasets for race: CMU Multi-PIE, MORPH, and RFW; and two datasets for age: Adience and CACD. LCNN: LightCNN; MSC: MS-Celeb-1M, VF2: VGG-Face2; RA: Race-A, RB: Race-B; A1: (0-14) years, A2: (15-32) years, A3: (33+) years.}
\label{tab:exp}
\begin{tabular}{|c|l|c|c|c|c|}\hline
\textbf{Exp.\#} & \textbf{Network} & \textbf{Pre-train} & \textbf{Train} & \textbf{Fine-tune} & \textbf{Test} \\ 
\hline
\hline
\multicolumn{6}{|c|}{\textbf{Effect of Race}}\\
\hline
\hline
1 / 2 & \multirow{3}{*}{LCNN-9} & \multirow{2}{*}{-} &  RA & - & RA / RB \\ 
\cline{1-1}
\cline{4-6}
3 / 4 &  &  & RB & - & RA / RB \\ 
\cline{1-1}
\cline{4-6}
5 /6 &  &  & RA, RB & - & RA / RB \\ 
\cline{1-6}
7 / 8 & \multirow{2}{*}{LCNN-29} & \multirow{2}{*}{MSC (10M)} &  - & - & RA / RB\\ 
\cline{1-1}
\cline{4-6}
9-12 & & &   & RA / RB & RA / RB\\
\cline{1-6}
13 / 14 & \multirow{2}{*}{ResNet50} & \multirow{2}{*}{MSC (10M),} & - & - & RA / RB \\
\cline{1-1}
\cline{4-6}
15-18&  &  & - & RA / RB & RA / RB \\ 
\cline{1-2}
\cline{4-6}
19 / 20 & \multirow{2}{*}{SENet50} & \multirow{2}{*}{VF2 (3M)} & -  & - & RA / RB\\
\cline{1-1}
\cline{4-6}
21-24 &  &  & -  & RA / RB & RA / RB\\
\cline{1-6}

\hline
\hline
\multicolumn{6}{|c|}{\textbf{Effect of Age}}\\
\hline
\hline
25 & \multirow{3}{*}{LCNN-9} & \multirow{2}{*}{-} &  A1 & - &A1 \\ 
\cline{1-1}
\cline{4-6}
26 &  &  & A2 & - & A2 \\ 
 \cline{1-1}
\cline{4-6}
27 &  &  & A3 & - & A3 \\ 
\cline{1-6}
28-30 &  LCNN-29& MSC (10M)&  - & - &A1/A2/A3\\
\cline{1-6}
31-33 & ResNet50& MSC (10M), & -  & - &A1/A2/A3\\
\cline{1-2}
\cline{4-6}
34-36 & SENet50& VF2 (3M)& - & - &A1/A2/A3\\ 
\hline

\end{tabular}
\end{table}
\section{Does Deep Learning Encode Race-specific Information?}
\label{sec:ethnicity}
The \textit{own-race bias} in humans has been well established, where we are able to recognize people of our own race more easily as compared to individuals of other race \cite{raceFromFace,meissner01thirty}. Also referred to as the \textit{other-race bias} or \textit{cross-race effect}, it has extensively been studied in humans in the form of cognitive, behavioural, and neuro-imaging studies. Depending on the racial identity of the individual and the person to be identified, different facial regions have shown more contribution during the decision making process. In this study, we understand the behaviour of deep learning models to observe if they also focus on specific facial regions based on the race, as observed in humans.

\subsection{Are Deep Learning Networks Prejudiced?}
Experiments 1-6 (Table \ref{tab:exp}) are performed to analyze the presence of prejudice in deep learning networks via class activation maps and feature visualizations. LightCNN-9 models are trained from scratch on data belonging to different races. The aim is to understand if faces of different races are learned in a similar manner, or does the model utilize race-specific features for identifying individuals. 

\begin{figure}
\centering
\includegraphics[clip, trim={0.5cm 1cm 0.7cm 0cm}, width=3.2in]{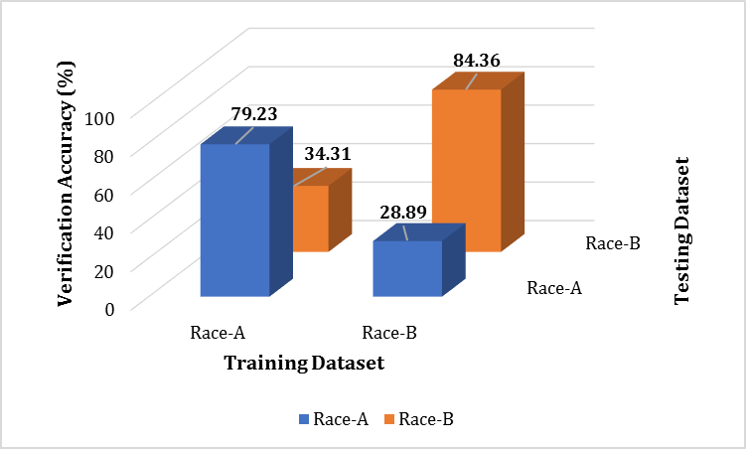}
\vspace{-10pt}
\caption{Verification accuracy (\%) at 1\% False Acceptance Rate of LightCNN-9 architecture. Models trained on images pertaining to a particular race demonstrate poor performance on face images of the other race.}
\label{fig:barGraphEthnicity}
\end{figure}

Figure \ref{fig:barGraphEthnicity} presents the GAR values obtained at 1\%FAR with LightCNN-9 models, for the two races. The model trained on face images of Race-A only reports an accuracy of 79.2\% on the testing set of same race (Race-A). On the other hand, the model trained on Race-B faces, tested on Race-A individuals, reports a GAR of 28.9\%. This suggests the presence of race-specific prejudice being encoded in the network due to the variations in the input data. Similar trends are observed when testing race-specific models on data belonging to Race-B individuals only. The LightCNN-9 network trained on faces of Race-B subjects only reports a GAR of 84.3\% on samples of Race-B (same race), while the network trained on Race-A faces achieves a GAR of 34.3\% (different race). 

\begin{figure}
\centering
\includegraphics[width=3in]{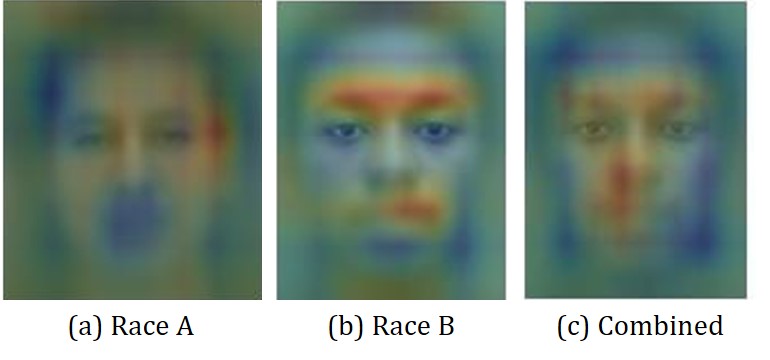}
\vspace{-10pt}
\caption{Visualization of salient regions obtained from exp. 1-5 (Table \ref{tab:exp}), where networks are trained from scratch on (a) only Race-A, (b) only Race-B, and (c) both Race-A and Race-B face images. Best viewed in color.}
\label{fig:lcnnEthnicity}
\vspace{-10pt}
\end{figure}

We also evaluated the models trained on a specific race on the RFW test set, and observed a similar trend, thus demonstrating that the drop in accuracy is caused due to the other-race bias and not dataset bias. When testing on the RFW Race-A test set, the network trained on Race-B faces in comparison to the network trained on Race-A, demonstrates a decrease of around 22\% of its performance. It is our belief that network prejudice occurs due to differences in the learned networks for varying input data.

Feature visualizations allow us to compare models beyond the verification accuracy, and analyze the regions being learned by the network to extract features with the highest discriminative information across races. Figure \ref{fig:lcnnEthnicity}(a-b) presents the salient regions used for feature extraction by the learned networks. These are obtained by interpolating the final convolution layer filter responses and super-imposing on the input image. It can be observed that both models focus on different regions for feature extraction. Figure {\ref{fig:lcnnEthnicity}(c)}  shows the visualizations of salient regions learned by a LCNN-9 model trained on equal Race-A and Race-B faces. The network learns a union of salient regions obtained for each race independently. While for the Race-B, salient regions correspond to the lips and above the eyes, Race-A appear to be distinguished on a more holistic view, with more focus on face boundary. \textit{The differences obtained in the feature maps, along with the accuracy variations strengthen the hypothesis that faces belonging to different race are encoded differently in a classification model, thereby suggesting presence of bias.}


\subsection{Do Deep Networks Mimic Humans?}
Cognitive studies in humans have analyzed how Race-A and Race-B individuals perform face recognition \cite{ellis75,peter06}. Across different studies, it has been observed that Race-B participants focus more on certain facial features such as mouth, lips, and nose while identifying other individuals of the same race \cite{ellis75}. On the other hand, Race-A subjects used traits such as the iris color, face shape, hair color and texture for describing and identifying other people of the same race \cite{ellis75,sinha19}. Such studies suggest a higher level of difference between the regions useful for distinguishing between individuals of different race. For example, information such as the eye color or hairstyle might not be useful for identifying a Race-B individual, whereas it could be of utmost importance for identifying an individual of Race-A.

\begin{figure}
\centering
\includegraphics[trim={0 1.49cm 0 0},clip,width=2.8in]{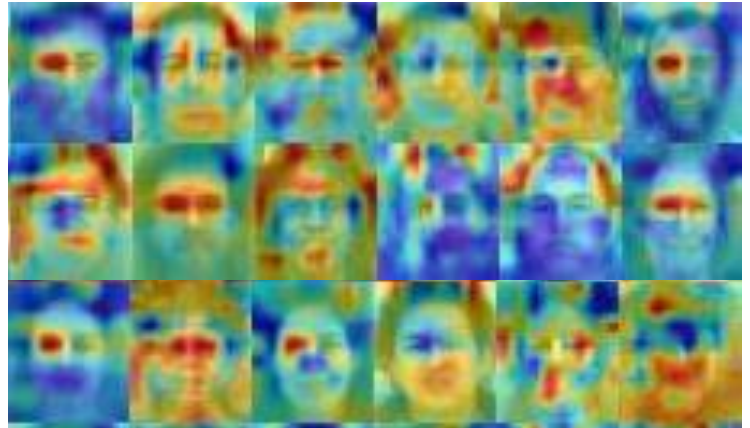}
\vspace{-10pt}
\caption{Sample average class activation maps obtained from the LightCNN-9 model trained on Race-A subjects. The model focuses heavily on the eye region for recognition. Best viewed in color.}
\label{fig:mpieCAM}
\vspace{-10pt}
\end{figure}

\begin{figure}
\centering
\includegraphics[trim={0 1.25cm 0 0},clip, width=2.8in]{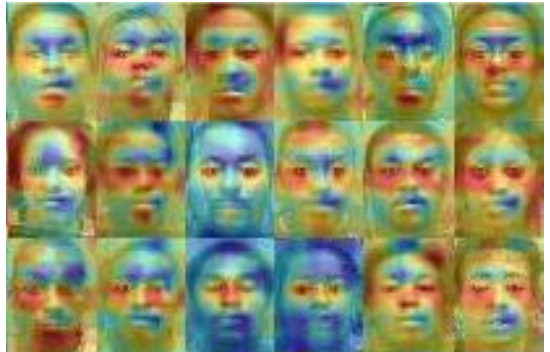}
\vspace{-10pt}
\caption{Sample average class activation maps obtained from the LightCNN-9 model on trained Race-B subjects. The model focuses heavily on the nose region (below the eyes), and the chin region. Best viewed in color.}
\label{fig:morphCAM}
\vspace{-10pt}
\end{figure}

Intrigued by the findings of human behavior, in this study, deep learning based face recognition systems are analyzed to investigate if they follow a similar pattern. In order to understand the behavior of deep learning networks, in terms of the useful regions of interest, class activation maps of the LightCNN-9 models trained on data pertaining to Race-A and Race-B for face recognition are analyzed (Experiment 1-6, Table \ref{tab:exp}). Networks trained on a particular race simulate the functioning of the human brain which has been in contact with individuals of a specific race only, thereby enabling us to understand own-race effect in deep networks. Figures \ref{fig:mpieCAM} and \ref{fig:morphCAM} present sample mean class activation maps. Each map corresponds to the mean activation associated with a particular class. It is interesting to note that the activation maps vary significantly between races, however, demonstrate a similar behavior within a particular race. The network trained on subjects belonging to Race-A only, identifies primarily on the basis of the eye region (Figure \ref{fig:mpieCAM}). These results are in conjunction with the cognitive studies reported in literature, where researchers have identified the eye color as one of the most identifiable traits used by humans as well \cite{ellis75}. \textit{This suggests that the behavior of deep learning networks and the human brain is alike.} Apart from the eyes, we also observe regions around the face being highlighted, which could mean face shape or hair type, both of which have been identified earlier from behavioural studies \cite{ellis75}. Similarly, Figure \ref{fig:morphCAM} presents sample mean class activation maps from the LightCNN-9 network trained on the Race-B dataset. Regions of the lip, nose, and cheekbones appear to contribute more towards face recognition, as compared to other facial regions. Similar results have also been reported in cognitive studies, wherein Race-B individuals utilized the eyes and lips for describing and recognizing other Race-B subjects \cite{ellis75}. \textit{The regions of interest used by models trained on specific races demonstrate the presence of \textit{different discriminative regions}, suggesting that networks \textit{sub-consciously} encode race-specific information, thereby resulting in an own-race bias.} 

\begin{figure*}
\centering
\includegraphics[width=6.8in]{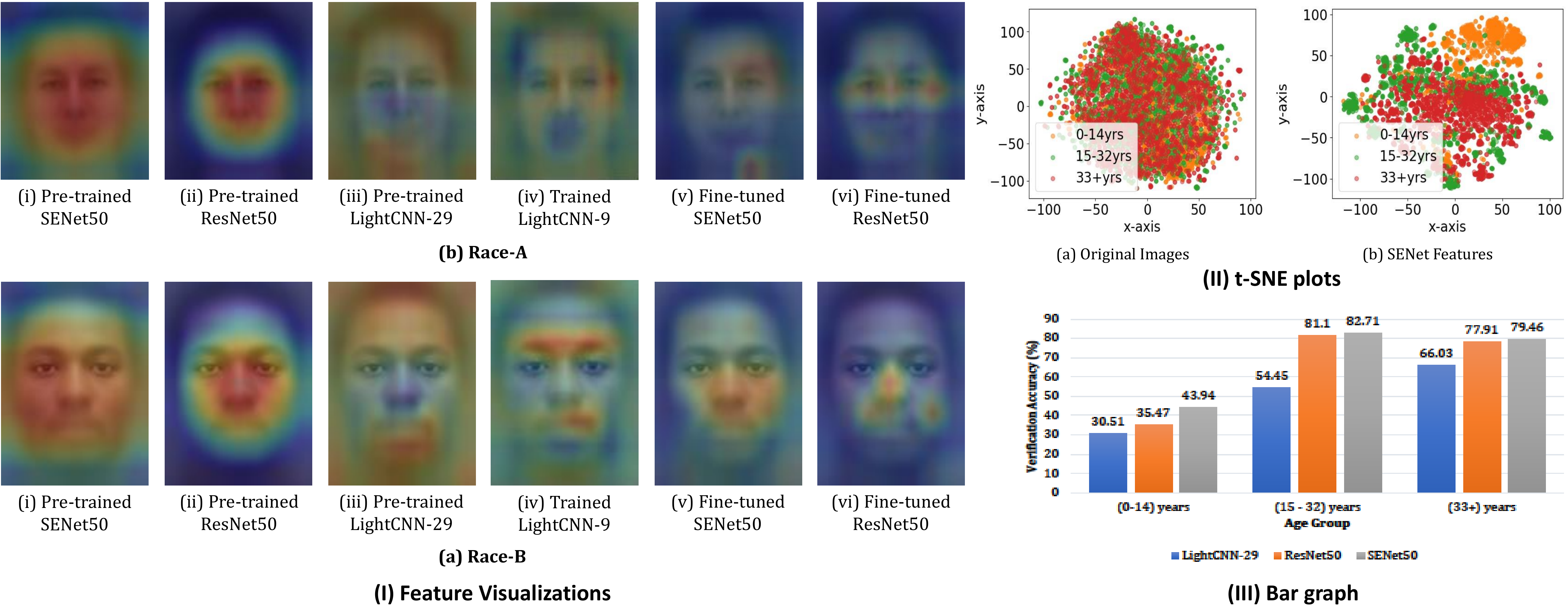}
\vspace{-10pt}
\caption{(I) Visualization of salient regions obtained from pre-trained, fine-tuned, and trained from scratch networks. Differences in the \textit{region of focus} are observed across models. (II) t-SNE visualizations of raw pixels and features extracted from SENet50. (III) Verification accuracy (\%) at 1\%FAR for three pre-trained networks on the three age based sub-groups. }
\vspace{-10pt}
\label{fig:featureEthnicity}
\end{figure*}

\subsection{More the Merrier: Does Large-scale Data Help in Mitigating Own-race Bias?}

The presence of own-race bias in individuals is often attributed to limited exposure to other race people \cite{lebrecht09perceptual}. In literature, studies have shown reduced effect of other-race bias upon training subjects to recognize face images of a different race. Since deep learning models rely heavily on the training samples, networks trained on large datasets may exhibit a similar behavior. In order to understand whether pre-trained networks suffer from the challenge of bias, deep networks trained on large-scale datasets are analyzed (Table \ref{tab:exp}: Exp. 7-24). The performance of three pre-trained networks - ResNet50, SENet50, and LightCNN-29 is studied. Feature maps and class activation maps are also computed to better understand their behaviour. Upon testing on the relatively constrained data belonging to Race-A, ResNet50 obtains an accuracy of 98.7\%, while for Race-B, the network attains a classification accuracy of 96.3\%. Similar results are obtained with the SENet50 network: 98.8\% and 96.5\%, respectively. The performance of LightCNN-29 network further depicts that the generalization of a network highly depends upon the training data used. It achieves an accuracy of 96.47\% and 78.12\% on Race-A and Race-B data, respectively. The results depict that pre-trained models are biased towards Race-A and achieve better performance on Race-A faces as compared to samples belonging to Race-B. 

Similar trend in results is observed on the unconstrained in-the-wild RFW dataset. The pre-trained ResNet-50 network achieves 77.3\% on Race-B and 90.6\% on Race-A, while the pre-trained SENet-50 model attains an accuracy of 76.4\% on Race-B versus 90.4\% on Race-A. Similar to the performance on the constrained data, the performance of the LCNN-29 model reduces to 58.2\% and 82.1\% for Race-B and Race-A, respectively. The consistent reduced performance on Race-B suggests that the generalization capability of a network depends heavily upon the amount and variability of the training data. Figure \ref{fig:featureEthnicity} presents the feature visualizations (region of interest) for the pre-trained networks. Both ResNet50 and SENet50 models learn feature maps which cover an entire circular region of the face, thereby using almost the entire face for recognition, which might result in higher generalizability.

Fine-tuning has emerged as a common practice for enhancing a network's performance for a chosen task and dataset. Experiments 9-12, 15-18, and 21-24 of Table \ref{tab:exp} are performed to understand the behaviour of deep networks after fine-tuning. It is interesting to note the transition in \textit{region of interest} from pre-trained networks to fine-tuned networks (Figure \ref{fig:featureEthnicity}). Initially, for both SENet50 and ResNet50, feature maps of both the races (Race-A and Race-B) are similar, and focus on a circular region covering almost the entire face. After fine-tuning, the updated feature maps appear similar to the ones discussed earlier (Figure \ref{fig:lcnnEthnicity}). The feature maps obtained after fine-tuning with images of Race-B are similar to the network trained only on Race-B, demonstrating a shift to only the nose region and face center. In case of Race-A individuals, the networks start focusing on the eye-region only. This demonstrates the importance of fine-tuning for specific problems. However, it also establishes that \textit{fine-tuning should only be performed when the test set is highly specific as it might reduce the generalization abilities of the network}.

\begin{figure*}
\centering
\includegraphics[trim={0cm 0cm 0cm 4cm}, clip, width=6.4in]{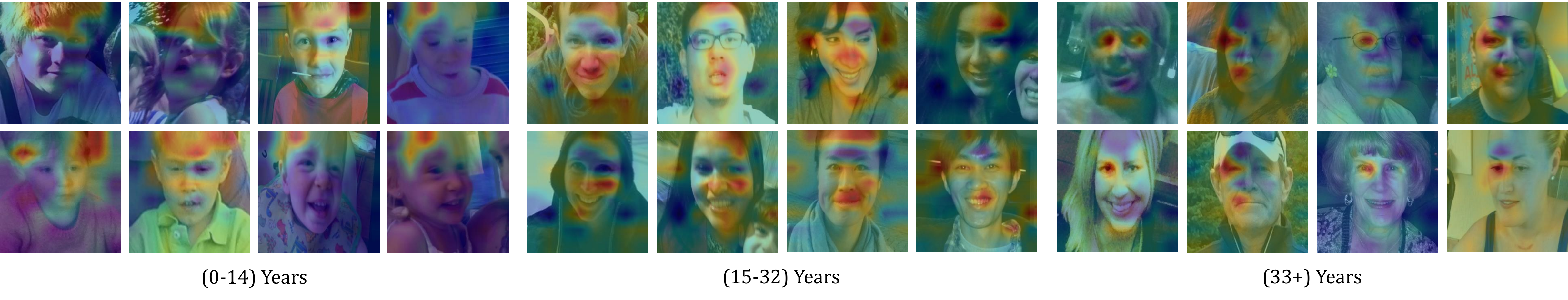}
\vspace{-10pt}
\caption{Class activation maps of sample images obtained from the trained LightCNN-9 networks. Best viewed in color.}
\label{fig:camAge}
\vspace{-10pt}
\end{figure*}

\section{Does Deep Learning Encode Age-specific Information?}
\label{sec:age}

Similar to the own-race bias, recently researchers have established the presence of \textit{own-age bias} in human beings \cite{anastasi05own}. Own-age bias refers to the phenomenon of an individual being able to identify other people of a similar/same age with a greater accuracy (and more easily) as compared to individuals of other ages. Research has focused on understanding the own-age bias on three age-groups: children, adults, and older people. While initial studies appeared conflicted in terms of an own-age bias in older people, however, across different studies, researchers have established its presence across different age groups \cite{wiese13ageing}. This is the first research which analyzes and studies the bias effect due to the factor of age in deep learning models. Previously, Kemelmacher-Shlizerman \textit{et. al} \cite{megaface} presented the MegaFace dataset and studied the impact of age on face recognition with respect to the data used for training a network. However, no analysis of deep models has been performed for the same. This section focuses on analyzing the own-age bias in deep learning models, specifically in terms of questioning its existence and investigating if distinguishing features vary across age. 

\subsection{How Well Do Deep Networks Recognize Children, Youngsters, and Adults? }
In order to analyze the behavior of deep learning networks in terms of own-age bias, a similar set of experiments are performed as the previous case study. Models (both trained from scratch (Table 1 \ref{tab:exp}: exp. 25-27) and pre-trained (Table \ref{tab:exp}: expt. 28-26)) are analyzed on face images of varying age-groups. The pre-trained networks of ResNet50, SENet50 and LightCNN-29 are evaluated for the three different age-groups: (i) 0-14 years, (ii) 15-32 years, (iii) 33+ years. The networks which are pre-trained on a large set of data capturing multiple variations, that is, ResNet50 and SENet50 obtained an accuracy of 77.9\% and 75.5\%, respectively for the oldest age group of (33+) years. Both the networks achieve a slightly higher performance on the age-group of (15-32) years, by reporting around 81-83\% (Figure \ref{fig:featureEthnicity}(III)), and less than 45\% for the youngest age group. On the other hand, LightCNN-29 reports a verification accuracy of 30.5\%, 54.45\%, and 66.03\% for the three age groups, in ascending order of age. \textit{The consistent lower face verification performance across different networks suggests the presence of an own-age bias in these networks.} Figure \ref{fig:featureEthnicity}(II) presents the t-SNE \cite{tsne} plots for the images of the test set (across all three age-groups), and the features extracted from SENet50. It is interesting to note that features of the youngest age-group (orange colour) appear to form a separate cluster, thereby suggesting that the network is able to distinguish between adults and children, however, the lower recognition performance suggest lower discrimination between the extracted features. The relatively improved performance of SENet50 and ResNet50 models (pre-trained on MS-Celeb-1M and VGGFace2) suggest that similar to the \textit{contact hypothesis} given in cognitive studies, where more exposure to a particular out-group results in increased recognition capabilities, more contact (training) with a sub-group of individuals might result in an improved classification performance for that sub-group. Thus, reinstating the importance of training data for modeling the different co-variates of face recognition. 


\subsection{What Does the Model See Based on the Age?}

Cognitive research has established the presence of own-age bias, however, limited research has focused on analyzing the specific regions of focus or discriminative facial features across different age groups. While there exist studies which analyze the gaze pattern or scan pattern of individuals belonging to different age-groups, they do not contribute much to the current study, since they do not specifically analyze the discriminative regions of interest. In this study, we analyze the behavior of face recognition models to investigate different regions of interest. 
 
Three LightCNN-9 models are trained, one for each age group, and corresponding class activation maps are presented in Figure \ref{fig:camAge}. The class activation maps demonstrate the discriminative regions of focus used by the model for classifying the input images. An interesting trend of activation is observed across the images of different age groups. \textit{Face images of kids (first four columns) appear to use more \textit{soft biometric} information such as hair and face shape for classification, while models trained on individuals of (15-32) years focus more on the lip, lower face, and forehead regions. Models trained on the third category, (33+) years rely heavily on the eye information for classification, along with other facial features.} The activation suggests the presence of different discriminative regions across age-groups, thereby suggesting the presence of an own-age bias, especially for young children below 15 years. The presence of different discriminative regions across varying age-groups provides new insights into the face recognition process, which opens the avenues for novel research directions in both deep learning and cognitive studies.

\begin{figure}
     \centering
     {\includegraphics[width=3.4in]{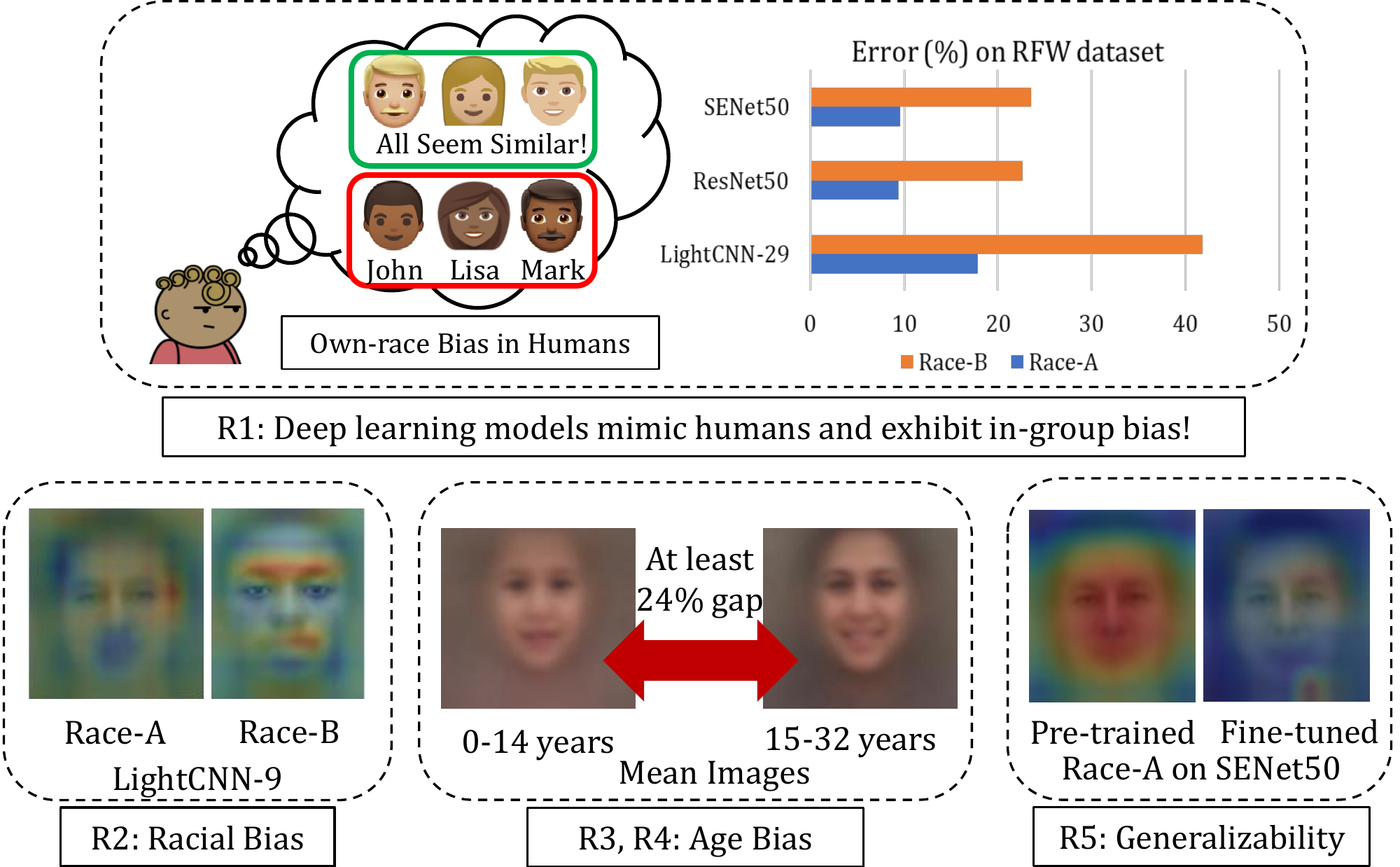}}
     \caption{Deep learning for face recognition- Prejudiced! }
     \label{fig:concl}
     \vspace{-20pt}
\end{figure}

\section{Conclusion}
Multiple instances have unfolded the \textit{hidden biases} of AI systems, creating a need for developing \textit{fairer systems}. In this direction, this research analyzes deep learning networks for the existence of bias in automated face recognition systems. This is the first work which attempts to answer \textit{if} and \textit{where} bias is encoded in face recognition systems, with respect to two case studies of race and age. Analysis across multiple deep learning networks suggests the presence of \textit{own-race} and \textit{own-age} bias in face recognition models. Similar to human behavior, deep learning networks demonstrate strong tendency of focusing on selected facial regions for a particular race, with variations across different race. To the best of our knowledge, this is the first study which suggests that the biased behavior of deep learning networks is similar to that observed in humans, therefore presenting opportunities of applying well established human cognitive results for bias elimination in deep learning networks. Our analysis also demonstrates the benefit of using large-scale data for training; however, does not recommend it as a complete solution for bias elimination. On the other hand, caution must be ensured while curating the training data for deep learning models, in order to incorporate maximum variability possible. Similar behavior is observed for the cross-age study, where face recognition networks appear to \textit{sub-consciously} encode the age information pertaining to the training samples resulting in varying regions of interest across groups, and presence of \textit{own-age bias}. Lower recognition performance of state-of-the-art face recognition systems on child face images further reaffirms the biased nature of such systems with respect to age as well. We believe that the insights provided in this paper can help facilitate research across multiple disciplines, and enable development of novel techniques for de-biasing existing face recognition systems.

{\small
\bibliographystyle{ieee}
\bibliography{egbib}

\begin{thebibliography}{10}\itemsep=-1pt

\bibitem{amazon}
Facial recognition, and bias.
\newblock \url{https://tinyurl.com/y7rat8vb}.

\bibitem{Google}
Google photos labels dark skinned people as gorillas.
\newblock \url{https://tinyurl.com/y7c7dxso}.

\bibitem{proPublica}
Propublica study of algorithms.
\newblock \url{https://tinyurl.com/gvtccpq}.

\bibitem{alvi18blind}
M.~Alvi, A.~Zisserman, and C.~Nellåker.
\newblock Turning a blind eye: Explicit removal of biases and variation from
  deep neural network embeddings.
\newblock In {\em European Conference of Computer Vision Workshops}, 2018.

\bibitem{anastasi05own}
J.~S. Anastasi and M.~G. Rhodes.
\newblock An own-age bias in face recognition for children and older adults.
\newblock {\em Psychonomic bulletin \& review}, 12(6):1043--1047, 2005.

\bibitem{anne2018women}
L.~Anne~Hendricks, K.~Burns, K.~Saenko, T.~Darrell, and A.~Rohrbach.
\newblock Women also snowboard: Overcoming bias in captioning models.
\newblock In {\em European Conference on Computer Vision}, pages 771--787,
  2018.

\bibitem{Bolukbasi16Nips}
T.~Bolukbasi, K.-W. Chang, J.~Zou, V.~Saligrama, and A.~Kalai.
\newblock Man is to computer programmer as woman is to homemaker? debiasing
  word embeddings.
\newblock In {\em International Conference on Neural Information Processing
  Systems}, pages 4356--4364, 2016.

\bibitem{genderShades}
J.~Buolamwini and T.~Gebru.
\newblock Gender shades: Intersectional accuracy disparities in commercial
  gender classification.
\newblock In {\em Conference on Fairness, Accountability and Transparency},
  volume~81 of {\em Proceedings of Machine Learning Research}, pages 77--91,
  2018.

\bibitem{vggface2}
Q.~Cao, L.~Shen, W.~Xie, O.~M. Parkhi, and A.~Zisserman.
\newblock Vggface2: A dataset for recognising faces across pose and age.
\newblock In {\em IEEE International Conference on Automatic Face \& Gesture
  Recognition}, pages 67--74, 2018.

\bibitem{cacd}
B.~{Chen}, C.~{Chen}, and W.~H. {Hsu}.
\newblock Face recognition and retrieval using cross-age reference coding with
  cross-age celebrity dataset.
\newblock {\em IEEE Transactions on Multimedia}, 17(6):804--815, 2015.

\bibitem{das18mitigating}
A.~Das, A.~Dantcheva, and F.~Bremond.
\newblock Mitigating bias in gender, age and ethnicity classification: a
  multi-task convolution neural network approach.
\newblock In {\em European Conference of Computer Vision Workshops}, 2018.

\bibitem{adience}
E.~Eidinger, R.~Enbar, and T.~Hassner.
\newblock Age and gender estimation of unfiltered faces.
\newblock {\em IEEE Transactions on Information Forensics and Security},
  9(12):2170--2179, 2014.

\bibitem{ellis75}
H.~D. Ellis, J.~B. Deregowski, and J.~W. Shepherd.
\newblock Descriptions of white and black faces by white and black subjects.
\newblock {\em International Journal of Psychology}, 10(2):119--123, 1975.

\bibitem{raceFromFace}
S.~Fu, H.~He, and Z.-G. Hou.
\newblock Learning race from face: A survey.
\newblock {\em IEEE Transactions on Pattern Analysis and Machine Intelligence},
  36(12):2483--2509, 2014.

\bibitem{multipie}
R.~Gross, I.~Matthews, J.~Cohn, T.~Kanade, and S.~Baker.
\newblock Multi-pie.
\newblock {\em Image and Vision Computing}, 28(5):807--813, 2010.

\bibitem{celeb}
Y.~Guo, L.~Zhang, Y.~Hu, X.~He, and J.~Gao.
\newblock {MS-Celeb-1M}: A dataset and benchmark for large-scale face
  recognition.
\newblock In {\em European Conference on Computer Vision}, pages 87--102, 2016.

\bibitem{resnet}
K.~He, X.~Zhang, S.~Ren, and J.~Sun.
\newblock Deep residual learning for image recognition.
\newblock In {\em IEEE Conference on Computer Vision and Pattern Recognition},
  pages 770--778, 2016.

\bibitem{peter06}
P.~J. Hills and M.~B. Lewis.
\newblock Reducing the own-race bias in face recognition by shifting attention.
\newblock {\em The Quarterly Journal of Experimental Psychology},
  59(6):996--1002, 2006.

\bibitem{senet}
J.~Hu, L.~Shen, and G.~Sun.
\newblock Squeeze-and-excitation networks.
\newblock In {\em IEEE Conference on Computer Vision and Pattern Recognition},
  2018.

\bibitem{tinyface}
P.~Hu and D.~Ramanan.
\newblock Finding tiny faces.
\newblock In {\em IEEE Conference on Computer Vision and Pattern Recognition},
  2017.

\bibitem{megaface}
I.~Kemelmacher-Shlizerman, S.~M. Seitz, D.~Miller, and E.~Brossard.
\newblock The megaface benchmark: 1 million faces for recognition at scale.
\newblock In {\em IEEE Conference on Computer Vision and Pattern Recognition},
  pages 4873--4882, 2016.

\bibitem{lfw}
E.~Learned-Miller, G.~B. Huang, A.~RoyChowdhury, H.~Li, and G.~Hua.
\newblock Labeled faces in the wild: A survey.
\newblock In {\em Advances in face detection and facial image analysis}, pages
  189--248. 2016.

\bibitem{lebrecht09perceptual}
S.~Lebrecht, L.~J. Pierce, M.~J. Tarr, and J.~W. Tanaka.
\newblock Perceptual other-race training reduces implicit racial bias.
\newblock {\em PloS one}, 4(1):e4215, 2009.

\bibitem{tsne}
L.~v.~d. Maaten and G.~Hinton.
\newblock Visualizing data using t-sne.
\newblock {\em Journal of machine learning research}, 9:2579--2605, 2008.

\bibitem{story}
C.~Meissner and J.~Brigham.
\newblock Thirty years of investigating the own-race bias in memory for faces:
  A meta-analytic review.
\newblock {\em Psychology, Public Policy, and Law}, 7:3--35, 2001.

\bibitem{meissner01thirty}
C.~A. Meissner and J.~C. Brigham.
\newblock Thirty years of investigating the own-race bias in memory for faces:
  A meta-analytic review.
\newblock {\em Psychology, Public Policy, and Law}, 7(1), 2001.

\bibitem{biasGroups}
M.~M. Nicholls, O.~Churches, and T.~Loetscher.
\newblock Perception of an ambiguous figure is affected by own-age social
  biases.
\newblock {\em Scientific Reports}, 8, 12 2018.

\bibitem{aies}
I.~D. Raji and J.~Buolamwini.
\newblock Actionable auditing: Investigating the impact of publicly naming
  biased performance results of commercial ai products.
\newblock In {\em AAAI/ACM Conf. on AI Ethics and Society}, 2019.

\bibitem{hyperface}
R.~{Ranjan}, V.~M. {Patel}, and R.~{Chellappa}.
\newblock Hyperface: A deep multi-task learning framework for face detection,
  landmark localization, pose estimation, and gender recognition.
\newblock {\em IEEE Transactions on Pattern Analysis and Machine Intelligence},
  41(1):121--135, 2019.

\bibitem{morph}
A.~W. Rawls and K.~Ricanek.
\newblock {MORPH}: Development and optimization of a longitudinal age
  progression database.
\newblock In J.~Fierrez, J.~Ortega-Garcia, A.~Esposito, A.~Drygajlo, and
  M.~Faundez-Zanuy, editors, {\em Biometric ID Management and Multimodal
  Communication}, pages 17--24, 2009.

\bibitem{imdb}
R.~Rothe, R.~Timofte, and L.~V. Gool.
\newblock {DEX}: Deep expectation of apparent age from a single image.
\newblock In {\em IEEE International Conference on Computer Vision Workshops},
  2015.

\bibitem{ryu18inclusive}
H.~J. Ryu, H.~Adam, and M.~Mitchell.
\newblock Inclusivefacenet: Improving face attribute detection with race and
  gender diversity.
\newblock In {\em Workshop on Fairness, Accountability, and Transparency in
  Machine Learning}, 2018.

\bibitem{sinha19}
P.~Sinha, B.~Balas, Y.~Ostrovsky, and R.~Russell.
\newblock Face recognition by humans: Nineteen results all computer vision
  researchers should know about.
\newblock {\em Proceedings of the IEEE}, 94(11):1948--1962, 2006.

\bibitem{rfw}
M.~Wang, W.~Deng, J.~Hu, J.~Peng, X.~Tao, and Y.~Huang.
\newblock Racial faces in-the-wild: Reducing racial bias by deep unsupervised
  domain adaptation.
\newblock {\em CoRR}, abs/1812.00194, 2018.

\bibitem{wiese13ageing}
H.~Wiese, J.~Komes, and S.~R. Schweinberger.
\newblock Ageing faces in ageing minds: a review on the own-age bias in face
  recognition.
\newblock {\em Visual Cognition}, 21(9-10):1337--1363, 2013.

\bibitem{lightcnn}
X.~Wu, R.~He, Z.~Sun, and T.~Tan.
\newblock A light {CNN} for deep face representation with noisy labels.
\newblock {\em IEEE Transactions on Information Forensics and Security},
  13(11):2884--2896, 2018.

\bibitem{cam}
B.~Zhou, A.~Khosla, A.~Lapedriza, A.~Oliva, and A.~Torralba.
\newblock Learning deep features for discriminative localization.
\newblock In {\em IEEE Conference on Computer Vision and Pattern Recognition},
  2016.

\end{thebibliography}
}

\end{document}